\documentclass[conference]{IEEEtran}

\usepackage{amsmath}
\usepackage{subfigure}
\usepackage{graphicx}
\usepackage{indentfirst}
\usepackage{latexsym}
\usepackage{dsfont}
\usepackage{txfonts}
\usepackage{amsfonts}
\usepackage{amssymb}

\ifCLASSINFOpdf
\else
\fi
\begin{document}
%
\title{Adaptive Training of Random Mapping for Data Quantization}

%

\author{\IEEEauthorblockN{Miao Cheng}  
\IEEEauthorblockA{
College of Information Engineering\\
Qingdao University\\
Qingdao, China\\
E-mail: mew\_cheng@outlook.com
}

\and
\IEEEauthorblockN{Ah Chung Tsoi}
\IEEEauthorblockA{Faculty of Information Technology\\
Macau University of Science and Technology\\
Macau S.A.R.\\
E-mail: actsoi@must.edu.cn}}


%


\maketitle

\begin{abstract}
Data quantization learns encoding results of data with certain requirements, and provides a broad perspective of many real-world applications to data handling. Nevertheless, the results of encoder is usually limited to multivariate inputs with the random mapping, and side information of binary codes are hardly to mostly depict the original data patterns as possible. In the literature, cosine based random quantization has attracted much attentions due to its intrinsic bounded results. Nevertheless, it usually suffers from the uncertain outputs, and information of original data fails to be fully preserved in the reduced codes. In this work, a novel binary embedding method, termed adaptive training quantization (ATQ), is proposed to learn the ideal transform of random encoder, where the limitation of cosine random mapping is tackled. As an adaptive learning idea, the reduced mapping is adaptively calculated with idea of data group, while the bias of random transform is to be improved to hold most matching information. Experimental results show that the proposed method is able to obtain outstanding performance compared with other random quantization methods.
\end{abstract}

%

%

\section{Introduction}
Many information processing problems refers to efficient data representation as a basic procedure of pattern analysis. As advancement in information technology, there emerges much demands on useful handling tools of big data, and resulting solutions to data proceeding. To fast retrieve required information, data quantization methods have been widely applied to find matched data with binary codes, and plays a key bridge between query and feedback in systems \cite{Andoni08neopha}\cite{Jegou11PQ}. It firstly learns the most outstanding features of each data, and then transforms the intrinsic patterns into binary codes as a global description. For each query, it is straightforward to find the best matched binary codes among data set with simple 'xor' comparisons, which usually temps to learn the data of similar category. And computational efficiency and processing performance are able to be calculated on, with such logical bits in comparison with other metric measures. In this work, the unsupervised learning based quantization is considered, which holds ubiquitously applicable advantages over other methods, due to the fact that no label information is required.

Till now, many the-state-of-arts unsupervised quantization solutions have been widely applied to data hashing. Among these methods, some of them learn the binary codes beginning with reduced generation of inputs for further binarization or data matching, as a basic step of random approaches. In the literature, many quantization method adopts well-known principal component analysis (PCA) \cite{Turk91PCA} to find the appropriate projections for binary embedding. As a popular hashing method, locality sensitive hashing (LSH) \cite{Datar04LSHpSD}\cite{Charikar02SER}\cite{Andoni08neopha} learns the low-dimensional representation of coming data with randomly generated projected directions. Spectral Hashing (SH) \cite{Weiss09SH} encodes input data into low-dimensional binary ones by preserving of local structures. And the similar idea is also used to find the best quantization with iterative procedures in other methods \cite{Gong11ITQ}. The fact is that, the resulting codes are always falling into binary formalism as belief results of cognitive contents. In other words, the obtained codes are of sequences of binary codes $ \left\{ 0,1 \right\}  $, whatever the range of inputs are given for 0-1 replacement. Furthermore, there needs a necessary threshold to constrain and filter inputs from original value extents, and positive / negative opinion has been logically employed as a natural deterministic decision on data distribution.

Furthermore, some specific methods, e.g., randomised-based hashing, has attracted more and more attentions in recent years as encoding features in randomised manners.
To address the binarization of arbitrarily given data, random function based hashing is devised to avoid fussy deduction and calculation burden of deterministic results \cite{Rahimi09RF}\cite{Raginsky09SIK}. And it is believed that, such random mapping holds an intrinsic relationship with Fourier features \cite{Lopez14RNCA}. In couple with inherence of binary codes by erasing periodic cycles of inputs, some works treat the hashing learning of data naturally with projected function values range in $ \left[ -1,1 \right]  $, which makes the bounding values naturally for binary encoders. Generally, one of its popular definitions can be referred as:
\begin{equation}
h\left( x \right) =sgn\left( f(x) \right) =sgn\left( cos\left( { w }^{ T }x+b \right)  \right),
\end{equation}
where $ sgn\left( \cdot \right) $ and $  cos \left( \cdot \right) $ respectively denotes the sign function and the cosine function, $ w $ denotes the
random mapping vector, while $ b $ denotes the linear offset \cite{Raginsky09SIK}. The inspired motivation behind such items can be referred to shift-invariant transform \cite{Lopez14RNCA}, e.g.,
\begin{equation}
 k\left( x,y \right) =k\left( x-y \right) =\int _{ { R }^{ d } }^{  }{ p\left( w \right) { e }^{ -i{ w }^{ T }\left( x-y \right)  }dw } .
\end{equation}
Then, the random features can be approximated to cosine formalism with well-known properties of functions of a complex variable,
\begin{equation}
\begin{matrix} cos & z \end{matrix}\equiv \quad \frac { { e }^{ iz }+{ e }^{ -iz } }{ 2 } .
\end{equation}
As a general outlet, the original shift-invariant kernels are to approximate the Gaussian kernel
$
 k\left( x,y \right) =exp\left( { -t\left\| x-y \right\|  }_{ 2 }^{ 2 } \right)
$
 with normal distribution of $ w $. Nevertheless, the binary coding of shift-invariant kernels are considered in this work for data quantization.

As a result, the mapped values of each data feature are calculated as
\begin{equation}
 x = {\left[ {cos\left( {w_1^Tx + b} \right),cos\left( {w_2^Tx + b} \right), \cdots, cos\left( {w_r^Tx + b} \right)} \right]^T} ,
\end{equation}
then the encoded data are obtained via the binary quantization.
Obviously, the final inputs to sign function is constrained by certain cosine function mapping values. In certain works, it is also regard as a specifical hashing function with sift-invariant mapping \cite{Raginsky09SIK}, if each feature is reformed into two angular transforms. For convenience, it is simply referred to cosine quantization (CQ) in this context, associated with respective transforms of each reduced features. There are also some existing works conducting the idea of randomized selection of data samples, which however, beyond the scope of this work.

Mostly, the original $ w $ and $ b $ are randomly generated in accordance with normal gaussian distribution as a common sense. Nevertheless, there is no existing discussion on alternative choices of random mapping, and optimal solutions to learn with binary balances are unavailable for further development. In this work, the adaptive trained quantization of random inputs will be discussed, and a variant of trigonometric function based quantization is proposed to learn the optimal mapping of initial inputs in light of 2-way clustering, while the offset rotations of random mapping are to be improved with maximum binary side information.

The following contents of this paper is organized as follows: section 2 will present the proposed improved hashing learning with adaptive cycle rotation, section 3 will evaluate performance of the proposed method with several random hashing methods. In section 4, we draw the conclusion of this work.

\section{Adaptive Training Quantization}
Supposed there given a set of data samples, $ X = \left[ {{x_1},{x_2}, \cdots, {x_n}} \right] \in {\mathds{R}^{d \times n}} $, ATQ aims to learn the linear mapping matrix $ W = \left[ {{w_1},{w_2}, \cdots,{w_r}} \right] \in {\mathds{R}^{d \times r}} $ and offset $ b $ to transform the original data into the $ r $-dimensional reduced features. As discussed foregoing, it is straightforward to randomly map the inputs into some low-dimensional space, and usually the ranged phase of results are out of ensured disciplinarian.

Actually, the outputs of angular based mapping function provides the transformed results between $ \left[ -1, 1 \right]  $, and naturally matches the finally reduced binary codes, e.g.,
$ cos \left( { w }^{ T }x+b \right) \in \left[ -1,1 \right] $.
However, the randomly selected $ w $ and $ b $ usually fall into a random result of data distributed range, and ideal quantization of cosine-like mapping is out of solution for further optimization. Here, a two-stage modified approach is devised to improve the resulting random mapping, so that more acceptable outputs are optimized to obtain the ideal quantization of CQ. To simplify depiction of discussions, $ Cos \left( \cdot \right) $ and $ Sin \left( \cdot \right) $ are employed to denote the resulting whole vector / matrix of cosine / sine transforms of each data feature, while $ cos \left( \cdot \right) $ and $ sin \left( \cdot \right) $ indicate the transforms of single input.


\subsection{Adaptive Training of Linear Mapping}
In terms of linear transform in pre-transform, the mapping is randomly generated for final binary coding. Nevertheless, it is considerable whether the two-side groups can be preserved in the reduced features as possible. If the class label information is ignored, the original problem may come down to a simple problem of 2-way groups. Without the supervised sense, the whole data is to be grouped into certain distribution with low self-cognitive nature in reduced space. In the literature, PCA has been widely adopted to seek for principal components of global distribution of data, and is able to preserve the main information of original data \cite{Turk91PCA}. In addition, it is also competent to learn the ideal data groups of feature structures as a data decomposition method, and informative contents of data models can be preserved.

To address the adaptive training of linear transform, the 2-way groups based objective function is devised to improve ideal mapping of original transform. Here, the offset of linear mapping is ignored firstly, so that handling of the first stage could be smooth, however, it is to be conducted in the following step. Without loss of generality, the objective can be defined as
\begin{equation}
J = \arg \mathop {\min }\limits_w Cos\left( {{w^T}X} \right) \Pi {\left( {Cos\left( {{w^T}X} \right)} \right)^T},
\end{equation}
where $ \Pi $ denotes the alignment matrix of data set, also known as Laplacian matrix. It is generally calculated as $ \Pi = I - \frac{1}{n}e{e^T} $, of which $ I \in \mathds{R}^{n \times n} $ denotes the identity matrix and $ e \in \mathds{R}^{n \times 1} $ denotes the vector with all '1' as its elements.

This objective is actually a kind of problems of data clusters in the principal subspaces, and has been widely applied to many self-taught applications. The original problem conducts the 2-way clustering of data, and seeks for solutions with principal decomposition of data \cite{Ding04KmPCA}. Nevertheless, it is different from the PCA based clustering methods with the general objective function, and traditional eigen-solution is unavailable for the further results. And also, there have been some works involving iterative approach to solve optimization problem of pattern analysis \cite{Quadrianto11MVNPP}\cite{Cheng14NCECA}. To optimize the involved objective, the Conjugate Gradient (CG) \cite{Bertsekas99NP} is adopted to seek for the local optimum for the general data mapping. The most advantages of CG method over other methods, e.g., Newton method \cite{Kelley99IMO}\cite{Bertsekas99NP}, is that it does not require to calculate the second-order gradient $ \nabla^2 J $, and optimized steps and directions are adaptively searched in iterations \cite{Cheng14NCECA}. More specifically, the conjugate direction of CG method is obtained by applying the Gram-Schmidt procedure to the gradient results iteratively, and step size is decided according to gradient values of the current and previous iterations.

For the referred objective function, its gradient can be calculated as
\begin{equation}
 \nabla J =  - 2X\left[ {Diag(Sin\left( {{w^T}X} \right))} \right] \Pi \left[ {Cos\left( {{w^T}X} \right)} \right].
\end{equation}
Here, $ Diag \left( \cdot \right) $ denotes the diagonal matrix with elements of input vector on its diagonal. By re-defining the $ t $-th gradient of objective function as $ g_t $, the CG method selects the suitable step size in the conjugate direction $ {s_t} \in {\mathds{R}^{d \times r}} $ with previously obtained conjugate direction $ s_{t-1} $, which is calculated as
\begin{equation}
 {s_t} =  - {g_t} + {\theta _{t - 1}}{s_{t - 1}}.
\end{equation}
Here, $ \theta _{t - 1} $
is computed based on gradients of original objective function $ g_t $ and $ g_{t - 1} $. In this work, the well-known Fletcher-Reeves formula is
adopted to calculate the $ \left( t - 1 \right) $-th $ \theta_{t-1} $, which is defined as
$ {\theta _{t - 1}} = {{{{\left\| {{g_t}} \right\|}^2}} \mathord{\left/
 {\vphantom {{{{\left\| {{g_t}} \right\|}^2}} {{{\left\| {{g_{t - 1}}} \right\|}^2}}}} \right.
 \kern-\nulldelimiterspace} {{{\left\| {{g_{t - 1}}} \right\|}^2}}} $.
Hence, $ W $ is updated as $ W{\rm{ = }}W{\rm{ + }}\alpha {s_t} $ with deduced step size $ \alpha $ in each iteration.

It is well-known that step size $ \alpha $ decides the reduction of the objective function with updated variable and in each iteration, and make the decrease sufficient. In the literature, a popular condition of sufficient decrease can be referred to the Armijo condition \cite{Bertsekas99NP},
\begin{equation}
 J\left( {{W^{\left( {t + 1} \right)}}} \right) - J\left( {{W^{\left( t \right)}}} \right) \le \lambda g_t^T\left( {{W^{\left( {t + 1} \right)}} - {W^{\left( t \right)}}} \right),
\end{equation}
where $ W^{\left(t \right)} $ denotes the obtained $ W $ in the $ t $-th optimization, and $ 0 < \lambda  < 1 $. On the other hand, the stopping rule plays an important role in iterative optimization problems, and an appropriate one is able to reduce overall computational cost. For general objective optimization problems, though the global minimum is difficult to achieve, it is feasible for the limit points to be stationary, while the local minimum can be found. In the literature, gradient based stopping rule has been a successful tool for stopping iterative search, it is mainly applied to quadratic optimization, while diseased learning problem is usually unavoidable in general objective functions \cite{Cheng14NCECA}.

To tackle the general objective, the numerical stopping criterion \cite{More91ns}\cite{Cheng14NCECA}\cite{Zdunek07nmf} is applied in this work, which is found much more effective than standard gradient based ones and diseased learning problem can be avoidable \cite{Cheng14NCECA}. With the values of objective function, it can be defined as
\begin{equation}
 J\left( {{W^{\left( t \right)}}} \right) - J\left( {{W^{\left( {t + 1} \right)}}} \right) \epsilon \mathop {\max }\limits_{k < t} J\left( {{W^{\left( k \right)}}} \right) - J\left( {{W^{\left( {k + 1} \right)}}} \right),
\end{equation}
where $ 0 < \epsilon < 1 $ denotes the constant of stopping rule. Obviously, the above stopping criterion focuses on the reduction of objective function value, while no gradient is needed to decide the stopping of optimization. Compared with gradient based stopping rule, it is more component for the iterative optimization of general objective function. In addition, further improved efficiency can be attainable if simple calculation is adopted for this stage, if no specific demand is required. And an alternative choice could be reached by using the randomized nonlinear component analysis methods. But the two-way group idea is considered in this work to demonstrate the possibility of adaptive training of random mapping.

\subsection{Adaptive Training of Offset}
In this subsection, the choice of offset constant of linear transform in cosine random mapping is to be discussed, so that appropriate rotation of angular is possible for acceptable results of binary embedding. To the best knowledge of ours, there has been no existing work conducting such topic, and a random constant is usually referred. Indeed, the offset constant plays a weak impact on final quantization as usual concept. Nevertheless, the randomly selected offset actually cannot match the optimal projective direction with cycle rounding generally as illustrated in Fig. 1.
\begin{figure}[htb]
  \centering
  \centerline{\includegraphics[width= 8.5 cm]{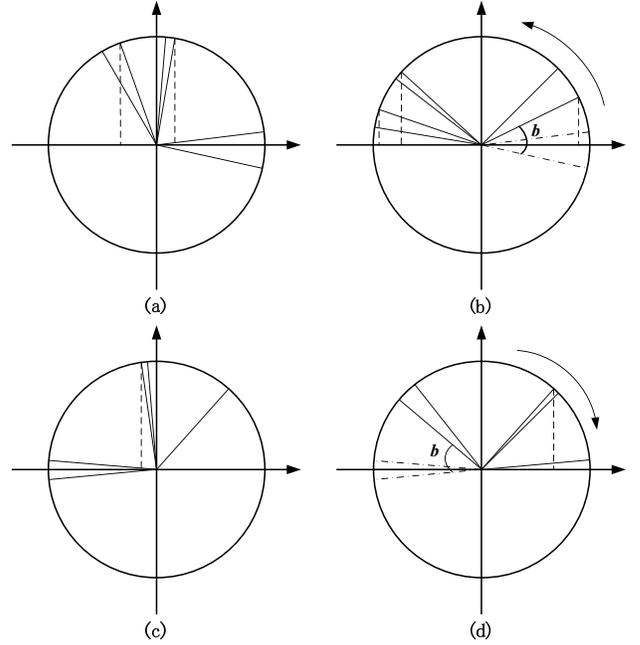}}
  \caption{The basic idea of adaptive training of offset. (a) and (c): The original mapping results of linear transform. (b) and (d): The adaptive rotation of biases.}
\label{fig:res}
\end{figure}

As the original bias of cosine function is determined by randomly selected offsets, the final hashing outputs may lay to an unbalanced hyperplane as a result. Then, the resulting mapping fails to maximally depict the projective values in the binary embedding as possible. In other words, the random biases are unable to make an optimal side decision of binary information for input data. This problem has been an intrinsic limitation of random hashing, and intuitively, the projective extension is required to be enlarged on horizontal direction as deterministic cosine hashing with maximum probability. This motivation has inspired this work as an initial attempt at ideal choice of mapping bias, and devise an adaptive learning of offset decision for believable binary embedding.

Generally, there are several projective directions are selected for linear mapping, and each of them afford the representation of data patterns in one reduced dimension. In terms of this, the original bias of linear mapping can be adjusted in every projective hyperplane, and most distinctive binary codes can be obtained in the final thresholding. To discuss conveniently, a simple linear mapping function is referred here, and further more combination can be implementation following the similar deduction. With an optimally chosen offset $ b $ of linear mapping, the mapping function is defined as
\begin{equation}
 { f }\left( x \right) =\cos { \left( { w }^{ T }x + b \right)  } .
\end{equation}
As discussed, the ideal offset holds the efficiency of maximizing extension on projective directions. On the other hand, it only adjusts the results on every horizontal hyperplanes and pushes affective impacts for each linear mapping. That is, only the associated mapping results have the correspondence with the given offset.

With a given data set $  X=\left[ { x }_{ 1 }, { x }_{ 2 }, \cdots, { x }_{ n } \right ] \in { \mathds{R} }^{ d \times n } $ and rotated bias $ b $, it is feasible to consider the related mapping side information in a square sum of row results, e.g.,
\begin{equation}
J\left( b  \right) = arg\underset { b  }{ max } \sum _{ i=1 }^{ n }{ { \left( \cos { \left( { w }^{ T }{ x }_{ i }+ b  \right)  }  \right)  }^{ 2 } }
\end{equation}
With respect to such objective function, it is difficult to handle the square items, and reach available solution directly with
amounts of data. In order to optimize the above objective,
the objective function can be simplified by unfolding each item of the original one,
\begin{equation}
\begin{array}{ll}
J \left( b  \right) = & arg\underset { b  }{ max } \frac { 1 }{ 2 } \left[ \sum _{ i=1 }^{ n }{ \cos { \left( { 2w }^{ T }{ x }_{ i } \right)  }  }  \right] \cos { 2 b  } \\
    & -\frac { 1 }{ 2 } \left[ \sum _{ i=1 }^{ n }{ \sin { \left( { 2w }^{ T }{ x }_{ i } \right)  }  }  \right] \sin { 2 b  } +\frac { n }{ 2 }.
\end{array}
\end{equation}
By absorbing two summation sets of trigonometric functions of $ w^T x_i $ and removing halfway constant, it can be further capsulated as
\begin{equation}
 J\left( b  \right) =arg\underset { b  }{ max } \cos { \left( 2 b +\arctan { \mu  }  \right)  }.
\end{equation}
Here, $ \mu $ is the value of the ratio
\begin{equation}
\mu =\frac {  \sum _{ i=1 }^{ n }{ \sin { \left( 2 { w }^{ T }{ x }_{ i } \right)  }  }   }{  \sum _{ i=1 }^{ n }{ \cos { \left( 2 { w }^{ T }{ x }_{ i } \right)  }  }  }.
\end{equation}
Clearly, the ideal $ b $ is able to be calculated directly, and deterministic result is confidently assured.

In accordance with the reduced formalism of optimization, the ideal offset can be calculated as $ b = \left \lfloor - \frac { \mu  }{ 2 }   \right \rfloor  $ with cycle adjustment of tangent function. With the learned offset, the side information of hashing codes are able to be maximized matching with binary embedding of trigonometric function. Though such idea is pushed intuitively, the performed results are improved for most cases. The suspicion comes from the fact that two phases of adaptive learning aim to search reverse directions along total objective cost. Nevertheless, it is fine for quantization as chosed offset can only be modified with minor rotation of spherical angular, and actually, periodically repeated results is hardly to make definite conclusion with ideal exploitation of quantization of uncertain data.

\begin{figure}[htb]
\begin{minipage}[b]{1.0\linewidth}
  \centering
  \centerline{\includegraphics[width=8.5cm]{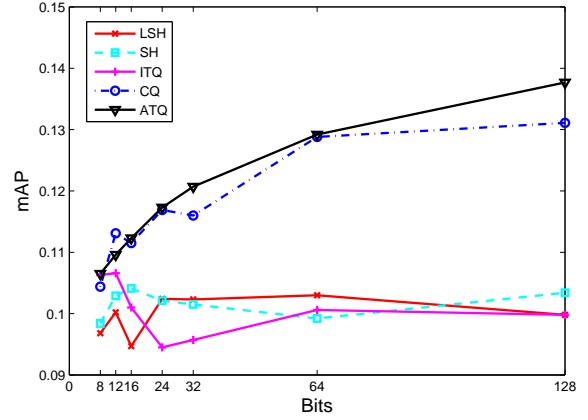}}
  \centerline{(a) Experimental results of CIFAR-10 data set.}\medskip
\end{minipage}
\begin{minipage}[b]{1.0\linewidth}
  \centering
  \centerline{\includegraphics[width = .99\linewidth]{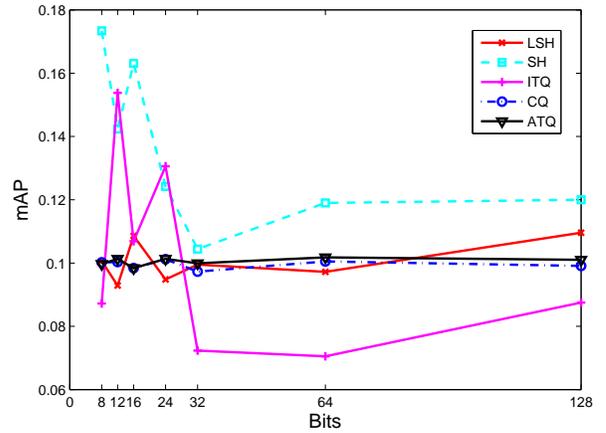}}
  \centerline{(b) Experimental results of MINIST data set.}\medskip
\end{minipage}
\caption{Experimental results against different code bits.}
\label{fig:res}
\end{figure}
\section{Simulation Evaluation}
In this section, the performance of proposed ATQ method is evaluated and compared with several the-state-of-arts random quantization methods for binary pattern matching, including LSH \cite{Andoni08neopha}\cite{Datar04LSHpSD}, SH \cite{Weiss09SH}, ITQ \cite{Gong11ITQ}, and CQ. Two data sets, CIFAR-10 image dataset \cite{Krizhevsky09cifar} and MINIST digit database \cite{Lecun98minist}, are used in these experiments. Furthermore, the mean average precision (mAP) is computed in each experiments to measure performance of different algorithms. In iterative learning of $ W $, the initial step $ \alpha $ is set to 1, and alterant step size $ \beta $ is set to 0.5. And the constants, $ \lambda $ of Armijo condition and $ \epsilon $ of stopping rule are set to 0.01.



For images in CIFAR-10 dataset, a set of 512 dimensional GIST descriptors \cite{Oliva01GIST} and 128 dimensional SIFT descriptors \cite{Lowe04sift} are learned from every tiny image, so that each data is represented as a 640 dimensional sample vector. In the experiments, the different batches in CIFAR-10 dataset is combined to one by ignoring their groups and several subsets are randomly selected to form the aforehand hashing data, query data and sequential data. For each data in MINIST, 784 dimensional gray features are used to describe its visual handwritten pattern. Similarly, the whole data are combined while the original order of data is disordered, and then lots of subsets are randomly selected to be involved into experiments.

In the experiment, the search performance of encoders are evaluated with two databases. For the CIFAR-10 data set, 10000 data are randomly selected and encoded with binary embedding, and then 1000 data are randomly selected to push forward matching query. For the MINIST data set, the training and testing data sets are combined into one set, respective 10000 and 1000 data are randomly selected to form the original and query data. It is noticeable that no supervised / semi-supervised corporation between them is involved, and only self-taught hashing is considered for adaptive learning. The first experiment evaluates the query performance of different algorithms with fixed 50 neighbors in mAP measure, and the search results with different hashing bits are shown in Fig. 1 (a) and (b).

According to the experimental results, the proposed ATQ and CQ are able to outperform all the-state-of-arts random quantization methods for the CIFAR-10 data set. And the obtained results of CQ and ATQ are quite similar to each other as increasing of encoding bits, and the tendency has been kept for large fluent data length. Other hashing methods give their results arounding mAP = 0.1, while involved data bits are used for binary embedding. LSH and SH can give better results compared with ITQ, if enough bits are used for query matching. And SH and ITQ present strong intensity of vibration with short bits, which is ameliorated with increased data bits.

For the MINIST data set, the results of most algorithms have vibrated strongly with different data bits. The similar dilemma also occurs in LSH, which also presents vibration for changing bits. Nevertheless, both CQ and ATQ always present stable results and the involved bits have no affection on their performance. Furthermore, the proposed ATQ method gives better results compared with original CQ, if limited binary bits are adopted to depict the data information.

\begin{figure}[htb]
\begin{minipage}[b]{1.0\linewidth}
  \centering
  \centerline{\includegraphics[width = .99\linewidth]{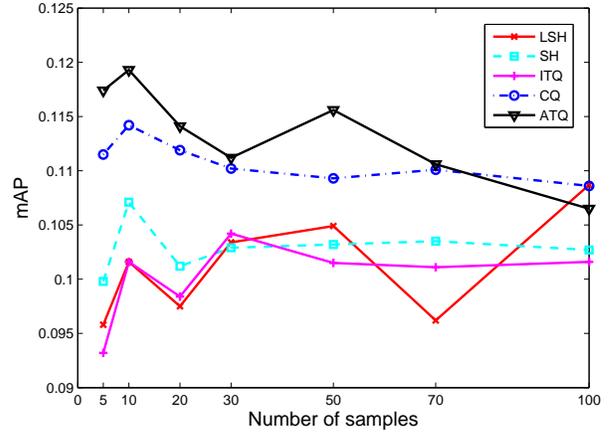}}
  \centerline{(a) Experimental results of CIFAR-10 data set.}\medskip
\end{minipage}
\begin{minipage}[b]{1.0\linewidth}
  \centering
  \centerline{\includegraphics[width = .99\linewidth]{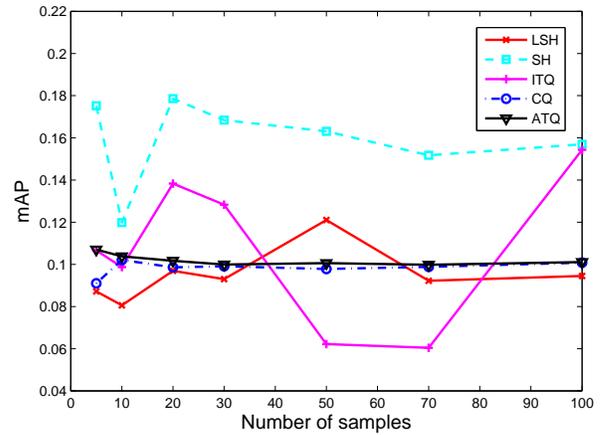}}
  \centerline{(b) Experimental results of MINIST data set.}\medskip
\end{minipage}
\caption{Experimental results against different number of samples used in mAP.}
\label{fig:res}
\end{figure}
Another experiment evaluates the performance of different algorithms with the obtained results against different number of samples used in mAP matching, and the results are shown in Fig. 2 (c) and (d). The experiments are proceeded while fixing the length of hashing bits and involved neighbors in mAP by turns. In details, different neighbors are used to calculate mAP, while 16 fixed bits are used to generate results.

Obviously, the results of different algorithms change with strong vibrations, as the involved samples in mAP is increasing stepwise. Nevertheless, SH and CQ can present much fluent results of stable performance compared with other methods. In terms of the results of CIFAR-10 data set, the ATQ and CQ work better than other methods in a higher rank, and ATQ takes the best results for most cases of different mAP. Compared with LSH and ITQ, SH gives much better results in most cases of different neighbors involved in matching query. And LSH presents the most fluctuant results with changing bits of quantization, among which best and worst performance is available.

On the other hand, the results from MINIST data set are quite different from the ones of CIFAR-10. SH presents the best performance though there is a large vibration for limited data samples used for calculation of mAP. The reason for this, can be considered as the reconstruction desire of MINIST data set with respect to intrinsic characteristics of SH. Furthermore, there are also large vibrations occurring in results of LSH and ITQ, following the ones of SH. Distinctively, CQ and ATQ are able to hold their results with stable performance with different number of involved samples in mAP. And it seems that, ATQ can give some better results over CQ if more neighbors are considered for data matching.

\section{Conclusion and Future Work}
As data quantization has been quite popular with data matching and efficient query, the related data encoding technologies have been studied broad in recent years. Specially, randomized quantization has been widely studied for binary hashing technologies, which usually considers the linear mapping as a random transformation. As a fascinating attempt, cosine-based mapping has been widely applied to find the binary codes of linear mapping, and it is also known as a variant of Fourier features if certain data unfolding is adopted. As a novel topic, it has attract much attentions with simple idea of binary embedding. Nevertheless, it usually confronts the limitation of uncertain projections, and obtained codes usually fails to fully describe the distinctive information of original data.

In this work, the limitation of devised cosine hashing is tackled, of which hashing is proceeded with random linear mapping. As the projective directions and offsets are generated randomly, there may exist unbalanced results in the low-dimensional results for further cosine thresholding.
As a result, an adaptive training of data quantization is proposed to improve the intrinsic limitation of cosine based encoder. In terms of linear mapping, two stage training approach is devised to learn the linear transformation and offset respectively.

In the first stage, the linear mapping is searched with the minimum objective of 2-way data groups. The fact is that, binary coding of data is
similar to data grouping problems, e.g., clustering, more or less. And PCA has been widely applied to find the group indicator of original data as
binary index if 2-way group is referred. In our work, such idea is adopted to construct the objective function so that self-organized data groups
are available under an unsupervised view.
For the offset of linear mapping, it presents a weak impact on cosine rotation of random mapping as a common sense. Nevertheless, it is disclosed that it can affect the final binary results as a adjustment of global results. And it is necessary to find an appropriate offset for improved results of binary embedding. In light of characteristics of cycle transformation, it is feasible to select the offset ideally so that binary side information could be preserved in the reduced codes. As a result, the obtained horizontal projections are able to preserve the maximum side information with the searched offset. According to experimental results, the proposed method is able to give an improvement on cosine random hashing, and output performance is derived with comparable results.

With respect to future work, it is valuable to exploit the latent performance of other trigonometric functions for encoder of linear mapping, and the computational efficiency is to be studied as a main issue. And the difference and combination of different functions are also a fascinating topic for further development on data unfolding based pattern analysis. Furthermore, the intrinsic relationship between trigonometric functions and frequency analysis of information is to be discussed with its applications to data processing. Based on this consideration, and latent solution to adaptive learning may be available for further study of data analysis.

\section*{Acknowledgment}
This work was supported by Natural Science Foundation Project of Macau S.A.R., research committee of Macau University of Science and Technology, and Qingdao University.
%



%

\bibliographystyle{IEEEtran}
\bibliography{ATQRef}

%
%

\end{document}